# Mixed Model OCR Training on Historical Latin Script for Out-of-the-Box Recognition and Finetuning

Christian Reul
Centre for Philology and Digitality
University of Würzburg
Germany

Christoph Wick
Planet AI GmbH
Rostock
Germany

Maximilian Nöth
Centre for Philology and Digitality
University of Würzburg
Germany

Andreas Büttner
Institute for Philosophy
University of Würzburg
Germany

Maximilian Wehner
Centre for Philology and Digitality
University of Würzburg
Germany

Uwe Springmann
CIS
LMU Munich
Germany

## ABSTRACT

In order to apply Optical Character Recognition (OCR) to historical printings of Latin script fully automatically, we report on our efforts to construct a widely-applicable polyfont recognition model yielding text with a Character Error Rate (CER) around 2% when applied out-of-the-box. Moreover, we show how this model can be further finetuned to specific classes of printings with little manual and computational effort. The mixed or polyfont model is trained on a wide variety of materials, in terms of age (from the 15$^{th}$ to the 19$^{th}$ century), typography (various types of Fraktur and Antiqua), and languages (among others, German, Latin, and French). To optimize the results we combined established techniques of OCR training like pretraining, data augmentation, and voting. In addition, we used various preprocessing methods to enrich the training data and obtain more robust models. We also implemented a two-stage approach which first trains on all available, considerably unbalanced data and then refines the output by training on a selected more balanced subset. Evaluations on 29 previously unseen books resulted in a CER of 1.73%, outperforming a widely used standard model with a CER of 2.84% by almost 40%. Training a more specialized model for some unseen Early Modern Latin books starting from our mixed model led to a CER of 1.47%, an improvement of up to 50% compared to training from scratch and up to 30% compared to training from the aforementioned standard model. Our new mixed model is made openly available to the community[1].

## 1 INTRODUCTION

Due to various large-scale book digitization efforts in the past a vast amount of printed material has become publicly available in the form of scanned page images. However, this can only be considered the first step in the effort to preserve and use our cultural heritage data. In order to make optimal use of these data it is necessary to convert the images into machine-readable and -actionable text enabling a host of use cases for researchers or the interested public.

Since large-scale manual transcription of the complete printing history is not feasible, we need to rely on automatic machine-learning methods (i.e., OCR) to recover the printed texts. For historical printings the straightforward application of OCR trained on modern fonts resulting in CERs of ca. 15% has been less than satisfactory, however, due to the multitude of peculiar fonts and character shapes used throughout history (see, e.g., [14, 18, 19]). Fortunately there have been major advances in this area, most notably by the introduction of a line-based recognition methodology based on LSTM networks [3] trained with the CTC algorithm [7], later further improved by the addition of convolutional neural networks (CNNs) [2]. This breakthrough enabled the high-quality recognition of even the earliest printed books by suitably trained models. However, in order to consistently achieve CERs below 2% or even 1% it is usually necessary to train *book-specific models* [10, 18] with Ground Truth (GT) consisting of pairs of line images and their corresponding transcription, whose creation is again a time consuming manual task.

To make the OCR of historical printings as comparably easy as the OCR of modern printings, one needs to construct *polyfont* or *mixed models* trained on a wide variety of different types, which are applicable out-of-the-box and without additional GT production and training. However, this approach usually leads to (considerably) higher CERs than using book-specific models [18]. While there are mixed models that consistently achieve CERs below 1% for e.g. Fraktur printings from the 19$^{th}$ century [13], this is much harder to realize for printings from the incunabula age due to the much more variant typography. In addition, whether a result is satisfactory highly depends on the requirements of the user: For preparing a critical edition of a book, even a CER of 1% may be too high for a tolerable manual correction effort. In contrast, in the context of mass OCR a higher CER might be acceptable as among other applications it enables full text search. To bridge the gap between these use cases one can build book-specific models starting from an existing (mixed) model [11] by *pretraining* and *finetuning*.

Therefore, the goal of this paper is twofold: First, we want to evaluate the construction and performance of a mixed model that can be applied to a wide variety of materials. Second, we want to show how the same model can be used as a starting point for a variety of training processes leading to more specialized models with minimal human and computational effort. As a concrete use case for finetuning we focus on Early Modern Latin printings from the 16th century, more precisely the works of the famous humanist Joachim Camerarius[2].

The remainder of the paper is structured as follows: After a brief overview of related work in Section 2 we explain the methodology of our experiments in Section 3 and introduce the data required for training and evaluation in Section 4. The experiments are described and discussed in Section 5 before we conclude the paper in Section 6.

## 2 RELATED WORK

Springmann et al. [17, 18] studied the effectiveness of mixed and book-specific models on (very) early printed books relying on the OCRopus OCR engine: First, they performed experiments on a corpus consisting of twelve books printed with Antiqua types between 1471 and 1686 with a focus (ten out of twelve) on early works produced before 1600. A two-fold cross evaluation yielded an average CER of the mixed model of 8.3%. Second, a similar experiment was conducted as part of a case study on the RIDGES corpus consisting of 20 German books printed between 1487 and 1870 in Fraktur. After applying the same methodology as mentioned above, the mixed models scored an average CER of 5.0%. Recognition with models trained for individual books, on the other hand, led to CERs of about 2% in both cases.

Breuel et al. [3] trained an OCRopus mixed model to recognize German Fraktur from the 19th century. The training data comprised around 20,000 mostly synthetically generated text lines which led to a model achieving CER of 0.15% and 1.37%, respectively on two books of different scan qualities.

Reul et al. [13] experimented with mixed model training on Fraktur data from the 19th century. The final Calamari[3] model was trained on over 250,000 lines of real GT from 150 sources. Since the GT was severely unbalanced between the sources (ranging from less than 50 lines to over 10,000 lines per book) a two-step training approach was employed which first used all available data and then refined the results by training on only a selected of up to 50 lines for each source, which showed to be highly effective. Evaluations on a variety of sources including novels, journals, and newspapers resulted in a very low error rate of 0.37%.

An approach of not only mixing different types but also various languages was promoted by [21]. They generated synthetic data for English, German, and French and used it for training language-specific models as well as a mixed one. As expected, the language-specific models performed best when applied to test data of the same language yielding CERs of 0.5% (English), 0.85% (German), and 1.1% (French). However, recognizing a mixed set of text data with the mixed models also led to a very low CER of 1.1% indicating a certain robustness of LSTM-based OCR engines regarding varying languages in mixed models.

We conclude that a construction and evaluation of the performance of mixed models constructed on and useful for a wide range of historical printings (age, language, types, etc.) is still missing.

## 3 METHODS

To achieve the best possible results we employ established techniques in the area of OCR training and deep learning in general such as pretraining, data augmentation and voting. We chose Calamari as our OCR engine since it is open-source, already showed state-of-the-art results in several areas of application [22], and provides GPU support for rapid training which is indispensable for our experiments (10 to 20 fold training time decrease compared to CPU training). In addition, Calamari supports many of the aforementioned techniques, including utilizing pretrained weights as a starting point for the training, the application of confidence-based voting ensembles [12], and data augmentation (DA) using Thomas Breuel's *ocrodeg* package[4]. Moreover, most recently the option to compute the Exponential Moving Average (EMA) of all weights similar to [15] has been introduced.

Regarding the data-related side of things we incorporated two independent variations into our training workflow. First, we adapted the composition of the data as proposed by [13], first training on all available data and later finetuning on a balanced set of GT for each contributing book. Second, we vary the input data by using different preprocessing results, i.e. mainly different binarization techniques, which can also be regarded as a form of data augmentation. In addition to selected solutions from the *ocrd-olena* package[5] [5] (Wolf, Sauvola MS Split), we also use the binary (*bin*) and normalized grayscale (*nrm*) images produced by OCRopus' *ocropus-nlbin* script [1]. Finally, we utilize the SBB binarization technique[6] of Staatsbibliothek zu Berlin based on [4] (*sbb*).

## 4 DATA

The GT for our experiments mostly consists of the original page image (color wherever available), derived preprocessing results, and the corresponding PAGE XML [9] file containing the line segmentation coordinates and their transcriptions. In addition, we constructed artificial page images from existing line based GT by vertically concatenating single line images and their corresponding transcriptions.

### 4.1 Training Data

Training data for our mixed model was taken from the following sources. By far the largest portion stems from the GT4HistOCR corpus [20] comprising over 310k lines of GT, available as binary and grayscale line images. About 80% belong to the *DTA19* subcorpus consisting of German Fraktur data from the 19th century. The remaining lines are a mixture of mainly German and Latin Fraktur and Antiqua data with a focus on the 15th and 16th century.

---

[2]http://kallimachos.uni-wuerzburg.de/camerarius
[3]https://github.com/Calamari-OCR/calamari
[4]cf. the *ocrodeg* package: https://github.com/NVlabs/ocrodeg
[5]https://github.com/OCR-D/ocrd_olena
[6]https://github.com/qurator-spk/sbb_binarization

Another large portion of the corpus consists of French Antiqua printings: The first part is the *OCR17* corpus [6] which has a strong focus on the 17[th] century but also contains data from the 16[th], 18[th], and 19[th] century. The second part was produced within the MiMo-Text project[7] of the University of Trier. It is based on double-keyed transcriptions of french novels published between 1750 and 1800.

The OCR-D [8] GT repository[8] comprises color images and PAGE XML files for 34 works from the 16[th] to 19[th] century with a strong focus on German Fraktur and a few Latin Antiqua printings.

As an additional resource for 19[th] Fraktur script we added further freely available resources like the Archiscribe corpus[9] and a repository for OCRopus Fraktur model training[10] as well as the evaluation data described in [13].

Finally, there is a considerable number of works collected from various projects and university courses using OCR4all [10] to produce GT. All pages are available as color images and there is a focus on pre-19[th] century German Fraktur printings. Among others, this includes a wide variety of images from the font dataset proposed by [16] which contains fonts that are underrepresented in the remaining data sets like Textura or Gotico-Antiqua. At the time of writing this paper, around 35 pages for each of the six selected font groups have been transcribed as a part of a project funded by the *Universitätsbund Würzburg* and the *Vogel foundation*.

As a case-study for finetuning our mixed model on some unseen books, we used the works of the well-known humanist Joachim Camerarius the Elder. These Early Modern Latin books were printed during the 16[th] century using various Antiqua types. For our experiments we selected five pages from each of the 19 books.

### 4.2 Evaluation Data

To evaluate our mixed models we used a set consisting of 29 historical books printed between 1506 and 1849. 16 books were contributed by the GEI Braunschweig[11]. The remaining 13 books were chosen to complement the GEI works in order to give a good representation of the relevant material with regards to age, print quality, fonts, etc. For each book we selected four to five pages and processed them using OCR4all: After performing the region segmentation with LAREX and the line segmentation with OCRopus we checked and, if necessary, corrected the line segmentation results before transcribing the text.

### 4.3 Transcription Guidelines

Before starting the training, we had to make several decisions regarding the set of characters for the final model, resulting in a set of guidelines for standardizing different transcriptions of the same glyphs in our ground truth. The most important rules are:

- Sticking to standard unicode and avoiding PUA (private use area) codes whenever possible
- Keeping vowel ligatures but resolving consonantal ones
- Regularizing all quotes to standard quotes and not differentiating between opening and closing ones

- Collapsing whitespace chains to a single one, removing whitespace at the start and end of a line, removing whitespace before punctuation and enforcing it after
- Regularizing different depictions of the same character as long as they do not carry semantic meaning (Umlauts, z/Z caudata, r rotunda, ...)
- Many minor regularization rules like mapping the capital letters *I* and *J* to *J*, sticking to simple double quotes as quotation marks, replacing macrons with tildes, etc.

## 5 EXPERIMENTS

All experiments were performed using Calamari version 2.1 and mainly its default parameters, most notably the default network structure of two CNNs, each followed by a max pooling layer and a concluding LSTM with a dropout ratio of 0.5. To ensure maximum comparability, we set the same random seed which standardizes all processes that involve randomness, for example data shuffling operations, data augmentation, and many more.

All trainings followed the default cross-fold-training methodology of [12] which produces five voters whose outputs are then combined by averaging their confidence values. Apart from the better recognition accuracy this has the advantage of considerably reducing the variance among the results, greatly improving comparability in the process.

The default early stopping methodology was applied: After each epoch the current model was evaluated against the held out validation data. Training was stopped if the validation CER did not improve for five consecutive times. In case no early stopping occurred, a training process lasted for a maximum of 100 epochs. Since there is a trade-off between training duration and recognition error rate, the number of samples after which an evaluation was done was set to half the size of the entire training set and the number of augmentations per sample was set to 5. Finally, we always utilized a general weight decay of $10^{-5}$ for all layers and an EMA weight decay of 0.99.

To evaluate a model we applied it to the evaluation data, standardized the output analogously to the training data, and calculated the CER using the Levenshtein distance between OCR result and GT.

### 5.1 Single-Stage Training

In this first experiment we compared different outcomes when only performing a single-stage training process. To get better insights into the influence of different parameters we varied the portion of the data used for training (all pages vs. only the selected ones), between applying data augmentation or not, and training on binary images alone (*bin*) or binary and grayscale combined (*ocro*)[12]. Table 2 sums up the results.

The first thing to notice is the generally low CER of around 2%, considering the variable and demanding material. As expected, the advantages of data augmentation are evident as all augmented models considerably outperforming their non-augmented counterparts. In general, models trained on all available data seem to perform slightly better than the ones that only see the selected pages. However, the main differences occur for the cases of training exclusively on bin and without augmentation (1 compared to 5) which seems sensible since for these cases the raw difference in training data cannot be compensated. Similarly, when using a comparatively small amount of training data and no augmentation, training on ocro improves the result by almost 9% (1 to 2). However, with more training data available (original or augmented) the difference becomes negligible.

---

[7]https://www.mimotext.uni-trier.de/en
[8]https://ocr-d.de/en/data.html#ocr-d-ground-truth-repository
[9]https://github.com/jbaiter/archiscribe-corpus
[10]https://github.com/jze/ocropus-model_fraktur
[11]Georg Eckert Institute for International Textbook Research, http://www.gei.de/en

[12]Since color images are not available for a large portion of the training pages we refrained from incorporating further binarizations in these first two experiments, since this would shift the composition of the data and make the results incomparable. For the Camerarius case study however, we made use of all available preprocessing outputs.

Table 1: Training and evaluation data used for our experiments. Apart from the *corpus* and *subcorpus* (if available) we list the *century, language* (German, Latin, French, and Dutch), the dominant *typography* class (Antiqua or Fraktur), the number of *works* and *pages* (all and selected ones), and whether *original* color page *images* are available.

| Corpus | Subcorpus | Century | Languages | Typography | # Works | # Pages all | # Pages sel | Original Images |
|---|---|---|---|---|---|---|---|---|
| GT4HistOCR | DTA19 | 19 | ger | f | 39 | 9,778 | 78 | |
| | EML | 15-17 | lat | a | 12 | 417 | 48 | |
| | ENHG | 15 | ger | f | 9 | 996 | 36 | |
| | Kallimachos | 15,16 | ger, lat, dutch | a, f | 9 | 840 | 36 | |
| | RIDGES | 15-19 | ger | f | 39 | 540 | 80 | |
| French | OCR17 | 16-20 | fr | a | 66 | 1,366 | 137 | ✓ |
| | MiMoText | 18 | fr | a | 40 | 3,315 | 20 | |
| OCR-D | - | 16-18 | ger, lat | a, f | 34 | 114 | 108 | ✓ |
| Fraktur19 | Archiscribe | 19 | ger | f | 103 | 165 | 165 | |
| | JZE | 19 | ger | f | 8 | 69 | 15 | |
| | Eval | 19 | ger | f | 19 | 188 | 38 | |
| Misc | - | 15-19 | ger, lat | a, f | 283 | 3,691 | 603 | ✓ |
| **Sum** | | | | | 642 | 21,479 | 1,364 | |
| Eval | GEI | 17-19 | ger, lat | a, f | 16 | 154 | | ✓ |
| | ZPD | 16-19 | ger, lat | f | 13 | 130 | | ✓ |
| **Sum** | | | | | 29 | 284 | | |
| Camerarius | Fold 1 | 16 | lat | a | 9 | 45 | | ✓ |
| | Fold 2 | 16 | lat | a | 10 | 50 | | ✓ |

Table 2: Results given as *CER* on binary images when performing a single-stage training process and varying the training *data* by using all or only selected pages (*part*), by utilizing different preprocessing results (*type*) and by incorporating data augmentation (*DA*) or not.

| ID | Data part | Data type | DA | CER (%) bin |
|---|---|---|---|---|
| 1 | sel | bin | | 2.10 |
| 2 | sel | ocro | | 1.92 |
| 3 | sel | bin | ✓ | 1.77 |
| 4 | sel | ocro | ✓ | 1.76 |
| 5 | all | bin | | 1.95 |
| 6 | all | ocro | | 1.95 |
| 7 | all | bin | ✓ | 1.80 |
| 8 | all | ocro | ✓ | 1.79 |

### 5.2 Two-Stage Training

To improve the results even further we applied the two-stage technique as proposed in [13]: After training on all available pages to show the model as many lines as possible, we used the resulting model as a starting point for the second training which was only performed on selected pages in order to balance the model. Again, we applied data augmentation and varied the image types. As a baseline we added a mixed model[13] trained on the entire GT4HistOCR corpus (no balanced second training step, no data augmentation) that is the standard Calamari model within the OCR-D project[14]. Table 3 shows the results.

As expected, the two-stage approach yielded better results than each of the single training processes from Table 2, since it combines the advantages of seeing all data, and consequently learning robust features (stage 1), and refining the model on a more balanced training set (stage 2). However, the gains are considerably smaller than expected, suggesting the assumption that the approach and the learning capacity of the model is about to saturate. Again, data augmentation improved the results, especially when applied during the first stage.

---
[13]https://qurator-data.de/calamari-models/GT4HistOCR
[14]https://ocr-d.de

Table 3: **Results given as *CER* on binary images when performing a two-stage training process and varying the training *data* by utilizing different preprocessing results (*type*) and by incorporating data augmentation (*DA*) or not.**

| ID | Stage 1 (all) | | Stage 2 (sel) | | CER (%) |
|---|---|---|---|---|---|
|  | type | DA | type | DA | bin |
| 9 | ocro |  | ocro |  | 1.80 |
| 10 | ocro |  | ocro | ✓ | 1.74 |
| 11 | ocro | ✓ | ocro |  | 1.73 |
| 12 | ocro | ✓ | ocro | ✓ | 1.74 |
|  |  |  | GT4HistOCR |  | 2.84 |

Table 4: **The ten most common confusions for the *GT4HistOCR* model and *LSH-4* showing the *GT*, prediction (*PRED*), the absolute number of occurrence of an error (*CNT*) and its fraction of total errors in percent (*PERC*).**

| GT4HistOCR | | | | LSH-4 | | | |
|---|---|---|---|---|---|---|---|
| GT | PRED | CNT | PERC | GT | PRED | CNT | PERC |
| ␣ |  | 1,156 | 20.96 | ␣ |  | 1,326 | 39.39 |
|  | ␣ | 881 | 15.98 |  | ␣ | 384 | 11.41 |
| ü | u | 62 | 1.12 | f | f | 36 | 1.07 |
| e | c | 53 | 0.96 | u | n | 23 | 0.69 |
| ö | o | 52 | 0.94 | Æ | E | 22 | 0.65 |
| , | / | 46 | 0.83 | c | e | 21 | 0.62 |
| f | f | 45 | 0.82 | n | u | 21 | 0.62 |
| J |  | 42 | 0.76 |  | . | 20 | 0.59 |
| W | V | 39 | 0.71 |  | , | 20 | 0.59 |
| ä | a | 38 | 0.69 | f | f | 19 | 0.56 |
|  | Remaining |  | 56.22 |  | Remaining |  | 43.79 |

The standard model on GT4HistOCR gets severely outperformed by all other models by close to 40%. While this naturally had to be expected due to the differences regarding the amount and balance of training data and partly the addition of augmentation, it still represents a very encouraging result with a view to the out-of-the-box application in mass OCR projects like OCR-D. For further experiments we from now on used model 11 as our best model and denoted it *LSH-4*[15].

### 5.3 Error Analysis

To get a better understanding of the nature of the achieved improvements we took a closer look at the most common errors of LSH-4 and compared it to the ones of the GT4HistOCR standard model (cf. Table 4).

The correct recognition of whitespaces represents by far the greatest amount of OCR errors, a well-known fact due to the irregular nature of inter-word spacings in Early Modern printings. Both the number of whitespace errors (2,037 to 1,710) and the number of substitution errors have been diminished in our new model.

---

[15]Latin Script Historical version 4. Available at https://github.com/Calamari-OCR/calamari_models_experimental

Table 5: **Results given as *CER* on OCRopus- (*bin*) and Staatsbibliothek-binarized (*sbb*) images when training on the Camerarius data starting from different models, namely from scratch, *GT4HistOCR*, and *LSH-4*, each with either training on OCRopus-binarized and grayscale (*ocro*) images or on all five available preprocessing variants (*all-var*).**

| ID | Start Model | Training | | CER (%) | |
|---|---|---|---|---|---|
|  |  | type | DA | bin | sbb |
| 1 | - | ocro |  | 2.98 | - |
| 2 |  | all-var |  | 2.42 | 2.17 |
| 3 |  | ocro | ✓ | 2.13 | - |
| 4 |  | all-var | ✓ | 1.74 | 1.60 |
| 5 | GT4HistOCR | - |  | 3.38 | 3.15 |
| 6 |  | ocro |  | 2.16 | - |
| 7 |  | all-var |  | 1.96 | 1.81 |
| 8 |  | ocro | ✓ | 2.07 | - |
| 9 |  | all-var | ✓ | 1.77 | 1.63 |
| 10 | LSH-4 | - |  | 2.22 | 2.32 |
| 11 |  | ocro |  | 1.50 | - |
| 12 |  | all-var |  | 1.43 | 1.40 |
| 13 |  | ocro | ✓ | 1.54 | - |
| 14 |  | all-var | ✓ | 1.47 | 1.36 |

Regarding the remaining errors, we would like to point out that the confusion between a comma (,) and a virgula (/) is spurious and is due to divergent transcription guidelines of different subcorpora of GT4HistOCR, where a printed virgula was sometimes transcribed as a slash and sometimes as a comma. Apart from that, both confusion tables mostly show somewhat different, yet quite common OCR errors like the confusion of similar characters (e/c, n/u, f/f, ...) or the insertions of spot-like punctuation marks probably caused by noise.

We conclude that the reduction of whitespace errors represents one of the greatest remaining challenges for models of historical OCR.

### 5.4 Finetuning Example: Camerarius

For this experiment we divided the Camerarius works into two sets consisting of nine and ten books, respectively, and performed a two-fold cross validation, always measuring the average of the two results. Apart from LSH-4 we again used the GT4HistOCR model for comparison and also compared against starting from scratch. As before, we incorporated data augmentation and varied the input data with regards to different preprocessing results (either binary images alone (*bin*), or all five available variants (*all-var*) combined). Since the data is already balanced and no refinement is necessary, we did not perform another two-stage training. Table 5 sums up the results.

The models produced by starting from LSH-4 outperform their GT4HistOCR and from scratch counterparts considerably in all scenarios, reaching excellent CERs around and under 1.5% on very challenging material. As expected, the degree of improvement declines with more training data, ranging from 50% when training on only *bin* from scratch without augmentation (row 1 compared to 11 in Table 3) to around 15% when training on all data types and incorporating augmentation (4 compared to 14). Interestingly, data augmentation does not improve the recognition when finetuning the LSH-4 model. This might be due to the fact that this model has already been augmented and is very well adapted to the material from its specific training.

Even though it is not the focus of our paper, it is worth mentioning that the *sbb* binarization looks very promising, as it enabled better recognition results[16] in all cases.

## 6 CONCLUSION

After collecting and preparing a large and multi-variant corpus consisting of over 21,000 pages, we were able to train a mixed model which outperforms a widely used standard model up to 40% when applied out-of-the-box to previously unseen material from 400 years of printing history. To achieve this, we enriched the training data even further by incorporating data augmentation, using different preprocessing outputs during the training process, and enforcing a two-stage training pipeline which first trains on all available data and then refines the resulting model on a more balanced selection.

The case study on the Camerarius works shows that using a general model as a starting point for a little finetuning with just one binarization and no data augmentation can reduce the CER by half (1.50% in this case compared to 2.98% when training from scratch). This is doable even without a GPU and within a few hours and therefore feasible for the practicing digital humanist.

Among the most promising areas for further advances for historical OCR we could identify the avoidance or postcorrection of whitespace errors (merges and splits of words). Other areas might be the use of color images, the incorporation of historical language models, and deeper or different neural network architectures.

Finally, despite there still being a lot to do, the fact that it is now possible to match results thought to be only achievable by book-specific training only a couple of years ago (cf. Section 2, especially [17, 18]), shows that historical OCR has come a long way in terms of training technology, techniques, and data.

---

[16]Application on the remaining binarization outputs (Wolf and Sauvola MS Split) led to considerably worse results compared to both *bin* and *sbb*.